\documentclass[lettersize,journal]{IEEEtran}
\usepackage{amsmath,amsfonts}
\usepackage{algorithmic}
\usepackage{algorithm}
\usepackage{array}
\usepackage[caption=false,font=normalsize,labelfont=sf,textfont=sf]{subfig}
\usepackage{textcomp}
\usepackage{stfloats}
\usepackage{url}
\usepackage{verbatim}
\usepackage{graphicx}
\usepackage{cite}

\usepackage{amsmath,amsfonts}
\usepackage{algorithmic}
\usepackage{algorithm}
\usepackage{array}
\usepackage{textcomp}
\usepackage{stfloats}
\usepackage{url}
\usepackage{verbatim}
\usepackage{graphicx}
\usepackage{color}
\usepackage{multirow}
\usepackage{arydshln}
\usepackage{bbding}
\usepackage{wrapfig}
\usepackage{booktabs}
\usepackage{colortbl}
\usepackage{newfloat}
\usepackage{listings}
\usepackage{arydshln} 
\usepackage{booktabs}
\usepackage{multirow}
\usepackage{bm}
\usepackage{bm}

\usepackage{booktabs}
\usepackage{threeparttable}

\usepackage{multicol}
\usepackage{multirow}

\usepackage{color} 
\usepackage{pifont}
\usepackage{mathrsfs} 
\usepackage{amsmath} 
\usepackage{colortbl} 
\usepackage{xcolor}   
\definecolor{lightblue}{RGB}{230, 242, 255}
\usepackage{bm}
\usepackage{makecell} 

\usepackage{algorithmic}
\usepackage{graphicx}
\usepackage{textcomp}
\usepackage{xcolor}
\def\BibTeX{{\rm B\kern-.05em{\sc i\kern-.025em b}\kern-.08em
    T\kern-.1667em\lower.7ex\hbox{E}\kern-.125emX}}
\newcommand{\KL}{D_{\mathrm{KL}}}
\newcommand{\CE}{\mathcal{L}_{\mathrm{CE}}}
    
\definecolor{cvprblue}{rgb}{0.21,0.49,0.74} 

\begin{document}

\title{Close Shortcut Wins Long: Seeking Diverse and Stable Generators for Data-Free Knowledge Distillation}

\author{
Kailin~Lyu,
Zherui~Zhang,
Junhao~Dong,
Kexue~Fu,
Weiguang~Pang,
Rongtao~Xu,
Qizheng~Wang,
Di~Wu,
Chee-Keong~Kwoh,
Longxiang~Gao,
Shibiao~Xu,
Changwei~Wang,
Ce~Hao,
and~Yu~Zhang%
\thanks{Kailin Lyu is with the Institute of Automation,
Chinese Academy of Sciences, and the Zhongguancun Academy, China.}%
\thanks{Zherui Zhang and Shibiao Xu are with the School of Artificial
Intelligence, Beijing University of Posts and Telecommunications, China.}%
\thanks{Junhao Dong and Chee-Keong Kwoh are with the School of Computer
Science and Engineering, Nanyang Technological University, Singapore.}%
\thanks{Kexue Fu, Weiguang Pang, Qizheng Wang, Longxiang Gao, and
Changwei Wang are with the Shandong Computer Science Center, China.}%
\thanks{Rongtao Xu is with Spatialtemporal AI, China.}%
\thanks{Di Wu is with the Institute of Automation,
Chinese Academy of Sciences, China.}%
\thanks{Ce Hao is with the Zhongguancun Academy, China.}%
\thanks{Yu Zhang is with Tongji University, China.}%
\thanks{Shibiao Xu and Changwei Wang are the corresponding authors.}%
}

\maketitle

\begin{abstract}
Data-Free Knowledge Distillation (DFKD) preserves privacy by transferring knowledge without real data access. However, existing generator-based DFKD methods suffer from over-reliance on teacher preferences and pattern collapse, exhibiting ``generative shortcut learning" in the frequency domain: dependent on specific frequency components and frequency positions, resulting in inconsistent synthetic image quality and class diversity. In this paper, we propose a \textbf{CSWL} framework aimed at introducing insights from the frequency domain perspective to improve generator diversity and training stability to \textbf{C}lose the phenomenon of \textbf{S}hortcut learning to \textbf{W}in in the \textbf{L}onger term. To address the issue of generative shortcut learning, we introduce frequency-domain augmentation at the feature level, encouraging the generator to attend to the full frequency spectrum and thereby suppress shortcut learning behavior. To tackle training instability, we propose a Cross-Stage Frequency Reconstruction (CSFR) auxiliary task, which implicitly constructs an Exponential Moving Average (EMA) mechanism to promote long-term optimization and stability. Extensive experiments, including downstream tasks and various image recognition datasets at multiple resolutions, validate the effectiveness of CSWL in improving both diversity and stability from the frequency view.
\end{abstract}


\section{Introduction}
\label{Introduction}

Knowledge distillation (KD) transfers knowledge from a large teacher network to a smaller student network, achieving similar performance with lower computational costs \cite{Guo2023,Klingner2023,Yuan2024}. Traditional KD methods assume student access to the teacher's training data, creating challenges in privacy-sensitive areas like healthcare, finance, and personalized services. Data-Free Knowledge Distillation (DFKD) \cite{Lopes2017,Fang2022} enables training without real datasets by reconstructing data from a pre-trained teacher network and particularly valuable in privacy-sensitive domains such as healthcare, finance, and personalized services. DFKD methods are divided into optimization-based \cite{Jiang2023,Zhao2023} and generator-based \cite{Wang2023a,Patel2023} approaches. This paper emphasizes generator-based methods, which employ adversarial generative training. Here, a pre-trained teacher network acts as a fixed discriminator, while a trainable generator produces proxy data mimicking the real dataset distribution, thus supporting knowledge distillation. In brief, DFKD proceeds in two iterative phases: data generation and knowledge transfer.
\begin{figure}[!t]
    \centering
    \includegraphics[width=0.9\linewidth]{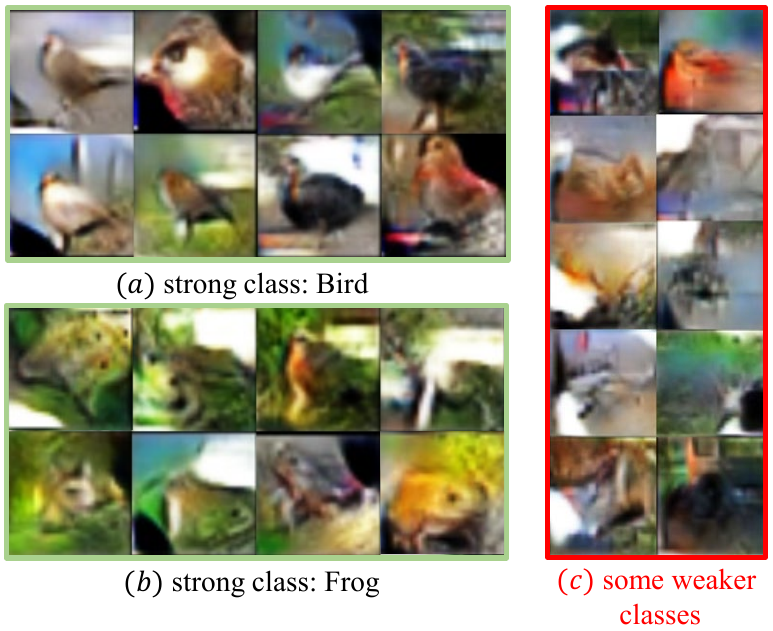}
    \caption{\textbf{Synthetic images were sampled from the mid-training phase of the generator (90 to 100 epochs).} Panels (a) and (b) illustrate the preferred categories, ``Bird" and ``Frog", which demonstrate relatively clear semantic content, whereas panel (c) displays images of weaker categories, characterized by abstract and noisy features.}
    \label{fig:mid-Synthetic images}
\end{figure}

\begin{figure*}[!t] 
    \centering 
    \includegraphics[width=0.95\textwidth]{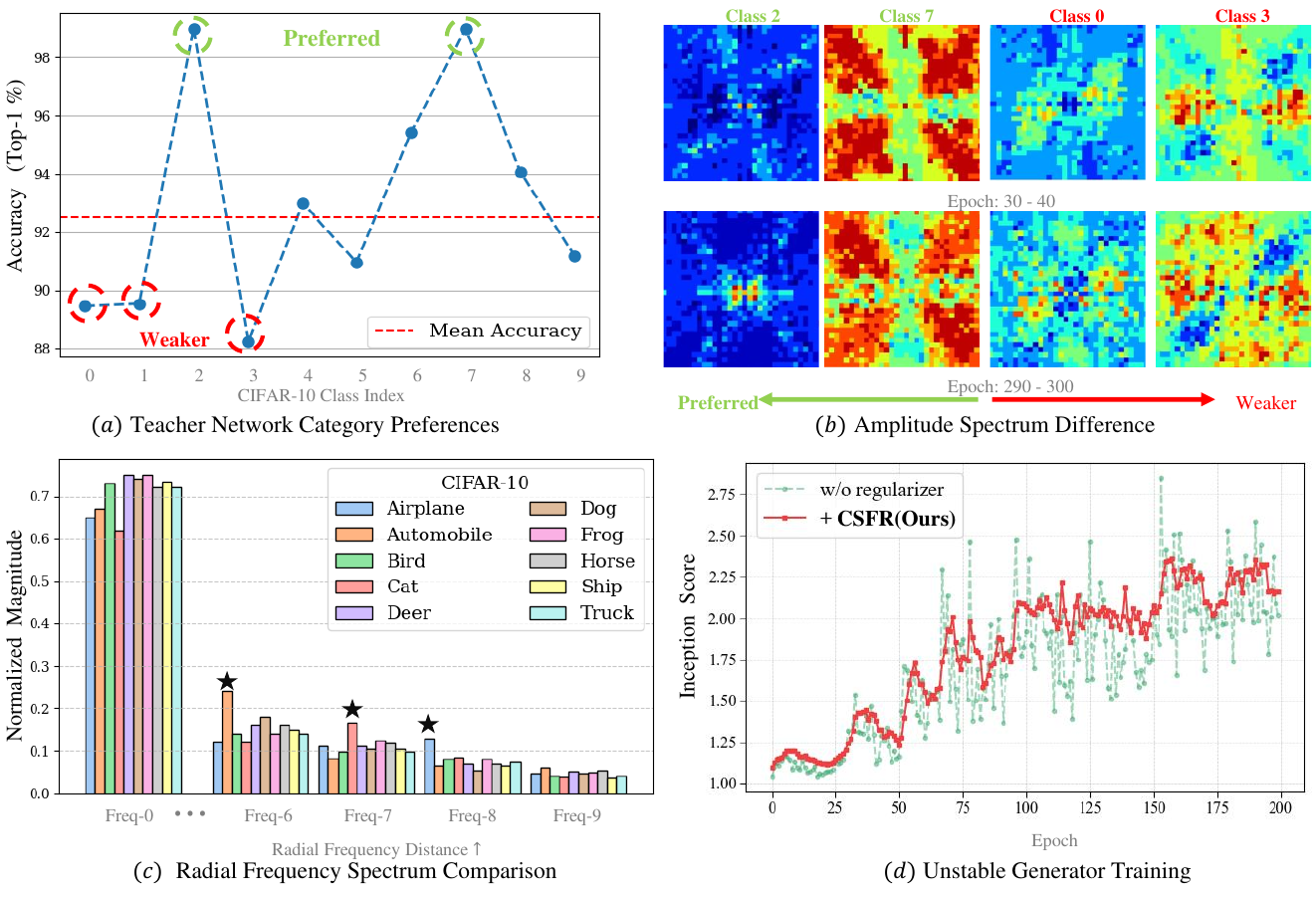} 
    \caption{\textbf{Frequency Domain Analysis of Generator-Based DFKD(e.g., the classic DAFD method).}
    (b) visualizes the ADCS, which measures the difference in the average Fourier amplitude of different classes at specific frequency locations. The red regions indicate that a particular class has significantly higher energy at the corresponding frequency, while the blue regions represent lower energy levels. 
    (c) presents the normalized magnitude distribution of CIFAR-10 classes across radial frequency ranges, with the x-axis denoting frequency bands (e.g., Freq-0 to Freq-9) and the y-axis representing the normalized magnitude. Higher magnitudes suggest a greater energy proportion at a specific frequency, whereas lower values indicate weaker dependence on it.
    (a) and (d) are introduced in the main paper.} 
    \label{fig:Frequency Domain Analysis1}
\end{figure*}

During data generation, synthetic images show quality, diversity gaps, and convergence inconsistency compared to real images across dataset categories. 
As shown in Figure~\ref{fig:mid-Synthetic images}, we sample synthetic images from the mid-training period (90-100 epochs) of the generator, focusing on preferred categories \textit{Bird} and \textit{Frog}. The preferred categories display relatively clear semantic content, while weaker categories remain abstract noise. Despite this, all images are used equally in the knowledge transfer phase, negatively impacting the student network's ability to learn balanced category knowledge. Existing efforts to improve category diversity focus on synthetic images (Image-Level) or generative features (Feature-Level), but these spatial strategies are abstract and hard for humans to interpret, complicating analysis \cite{Yu2023, Fang2021}.
To investigate the quality and diversity inconsistencies in synthetic image categories, We conduct previous DFKD methods on different datasets \cite{fang2019data,Yu2023,Fang2021} and explore the issue from a clearer frequency domain perspective, leading to these insights:

\noindent
\textbf{\textit{i).} Teacher Network Category Preference}: Figure~\ref{fig:Frequency Domain Analysis1}(a) shows that the teacher network exhibits varying classification accuracies across different categories in the CIFAR-10 dataset. As a result, the generator-based DFKD method receives inconsistent guidance from the teacher network during updates, exacerbating the imbalance in category-wise learning.

\noindent
\textbf{\textit{ii).} Generator Shortcut Learning:} CNNs' shortcut learning \cite{Wang2023a} relies on data biases, neglecting essential relationships. Similarly, Figure~\ref{fig:Frequency Domain Analysis1}(c) illustrates this in DFKD optimization, as all categories depend on low-frequency components (Freq-0) for semantic construction. Weakly predicted categories (\textcolor{red}{classes 0, 1, 3}) rely more on high-frequency ($\geq$ Freq-6) components than preferred (\textcolor{green}{2, 7}). This indicates that the generator's optimization process exhibits a strong dependency, making it prone to local optima. Consequently, this limitation hinders long-term optimization and significantly degrades the quality of synthesized data for certain categories.

\noindent
\textbf{\textit{iii).} Energy Accumulation:} Figure~\ref{fig:Frequency Domain Analysis1}(b) illustrates the distribution of Accumulative Difference of Class-wise average Spectrum (ADCS) \cite{Wang2023a} across frequencies. As the generator updates (epoch 30 $\rightarrow$ to epoch 290), preferred teachers like class 2 and class 7 show greater energy variability across all frequency positions compared to weaker classes like class 0 and class 3. This pattern indicates the generator's inherent bias toward certain categories. Furthermore, its optimization process reinforces these biases, reducing synthesized data diversity and compromising training stability.


Given the above insights, in this paper, we propose \textbf{CSWL} DFKD framework, which improving the class diversity or quality of synthetic images from a frequency domain perspective, consequently benefiting the student network. 
CSWL comprises two components: the \textbf{CDFA} module and the \textbf{CSFR} task.
Specifically, the CDFA module that decomposes intra-class amplitude and phase spectra, then performs efficient inter-class fusion to avoid the generator's dependence on specific frequency components and positions.
However, while indirectly introducing class diversity into the synthetic images, it is noteworthy that, as illustrated in Figure~\ref{fig:Frequency Domain Analysis1}(d), this improvement is unstable (with the IS score exhibiting significant fluctuations), which stems from the inherent conflict between the flat distribution required for class diversity and the sharp distribution required for knowledge transferability.
To address this instability, we propose a Cross-Stage Frequency Reconstruction (CSFR) Auxiliary task. CSFR masks and restores frequency responses containing significant amplitude information, thereby introducing an effective regularization mechanism into the generator optimization and encouraging the generator to focus more on the inherent class structure. In this work, our contributions can be summarized as follows:
\begin{itemize}
    \item \textbf{Frequency Domain Insights into DFKD.}
    Existing DFKD explorations primarily focus on the spatial domain, hindering human visual understanding and detailed analysis. We introduce a frequency domain perspective to reveal the causality between the generator's shortcut learning and the limitations in synthetic image quality and class diversity.
    \item \textbf{Dual Benefits in Diversity and Stability.} 
    Our proposed CSWL-DFKD framework comprises two components: CDFA and CSFR.  CDFA operates in the frequency domain, preventing reliance on specific frequency patterns and promoting DFKD class diversity.  CSFR, through a shared decoder paradigm, introduces regularization via salient information reconstruction, improving DFKD training stability.
    \item \textbf{Effectiveness}.  
    Extensive qualitative and quantitative experiments across diverse image recognition datasets at multiple resolutions, including downstream tasks, demonstrate the clear superiority of our frequency-domain-based CSWL framework.
\end{itemize}

\section{Related Work}
\label{Related Work}

\subsection{Knowledge Transfer}
With the rapid development of artificial intelligence \cite{my1,my2,my3,my4,my5}, traditional knowledge distillation transfers knowledge from a well-trained teacher model to a student model by applying temperature-scaled softmax to the teacher's logits and using the resulting softened probability distribution as a supervisory signal for training the student \cite{transfer1,transfer2}. Hinton et al. \cite{hinton} further proposed knowledge distillation, in which a smaller network is trained using the softened class probabilities produced by a larger pre trained teacher network. To explore the deeper knowledge structure of the teacher model, FitNets \cite{fitnets} employed L2 loss to align the intermediate feature maps of the teacher and student networks. Attention Transfer (AT) \cite{paying} introduced attention maps as a means of transferring knowledge, enabling the student network to better understand and imitate the behavior of the teacher network. However, pre trained models are sometimes released without the original training data because of privacy or storage constraints, which renders these methods inapplicable.

\subsection{Data-Free Knowledge Distillation} 
Data-Free Knowledge Distillation (DFKD) transfers knowledge from a pre-trained teacher network to a student network using synthetic datasets instead of original training data. Two key methods for generating these proxy datasets are optimization-based and generator-based approaches. Optimization-based methods Initially proposed by \cite{Wang2018,Zhao2020,Zhao2021}, these techniques iteratively update synthetic images with meta-learning. Various objectives have been explored, For example, Nayak et al. \cite{zero31} modeled the outputs of the teacher network as a Dirichlet distribution and used this distribution as a constraint during the optimization of noise images to produce synthetic samples. Yin et al. \cite{yin44} regularized the distribution of synthetic images using the statistical information stored in the batch normalization layers of the teacher network. Recently, generator-based methods using GANs have become popular, aligning student network predictions with teacher outputs using synthetic data \cite{Do2022,Binici2022}. CSD \cite{Luo2023} uses self-supervised tasks for adaptive data synthesis, while NAYER \cite{Tran2024} incorporates category priors from language models into the generator to speed up convergence. SpaceShipNet \cite{Yu2023} uses Generative Feature Exchange (GFE) to improve image diversity. Despite recent progress, challenges including ``shortcut learning" and insufficient pixel level exploration in the spatial domain persist, thereby limiting the performance of downstream tasks. To overcome these limitations, we propose a novel CSWL framework that enhances diversity and stability from the frequency domain perspective while enabling more intuitive qualitative and quantitative analyses.

\subsection{Learning in the Frequency Domain} 
Frequency domain analysis has attracted increasing attention in deep learning in recent years because of its unique advantages in image representation and generative model optimization \cite{Zheng2021,Cao2020}. The Fourier Transform (FT) converts signals to the frequency domain, revealing global features and structures, which are more effective than pixel-based features for capturing semantic information \cite{Zhang2024}. In knowledge distillation, frequency domain features have shown promise. Monsefi et al. \cite{Monsefi2024} proposed a frequency based generative module to enhance the diversity and quality of synthetic data, thereby providing more informative training samples for the subsequent distillation process. In addition, several studies have further revealed the distinct roles of different frequency components in generative tasks. For example, Li et al. \cite{Li2023} found that high-frequency components are crucial for generative model quality, while  \cite{Jung2023} combined an Adversarial Amplitude Generator with Dual Adversarial Training to improve neural network robustness. In this paper, we exploit the frequency domain to re-think the causality between the generator's reliance on teacher prediction preferences and the resulting limitations in image quality and diversity inherent in generator-based DFKD. Based on it, we provide a new analytical framework and theoretical insights into the intrinsic limitations of generator based data free knowledge distillation in the frequency domain.

\begin{figure*}[!ht] 
    \centering 
    \includegraphics[width=\textwidth]{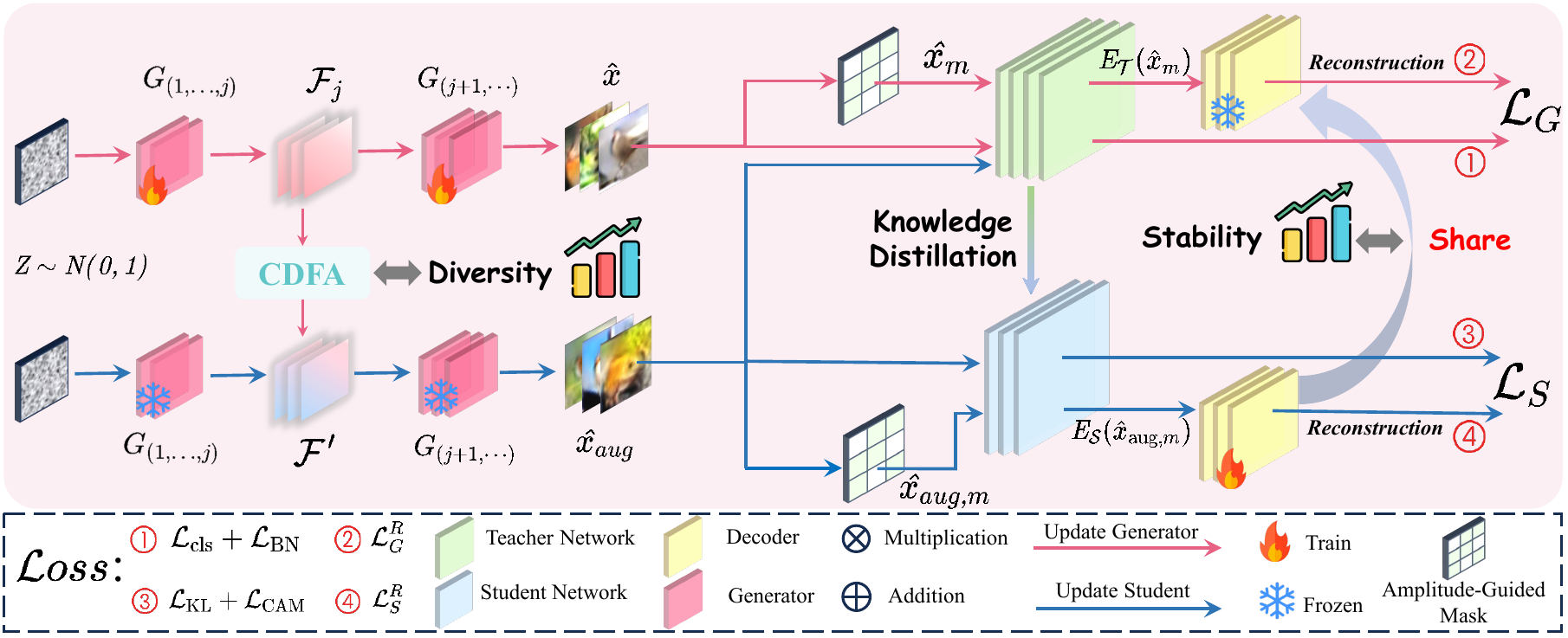}
    \caption{\textbf{Overview of the proposed framework.} The proposed framework employs a noisy layer and generator to synthesize images, enabling joint training of the generator, student network, and shared decoder without real data. Furthermore, CDFA and CSFR modules perform frequency-domain disentanglement and regularization, respectively, enhancing the \textbf{\textit{diversity}} and \textbf{\textit{stability}} of the generator while improving the generalization and representation transferability of the student network.} 
    \label{fig:overview} 
\end{figure*}

\begin{figure}[!t]
    \centering
    \includegraphics[width=\linewidth]{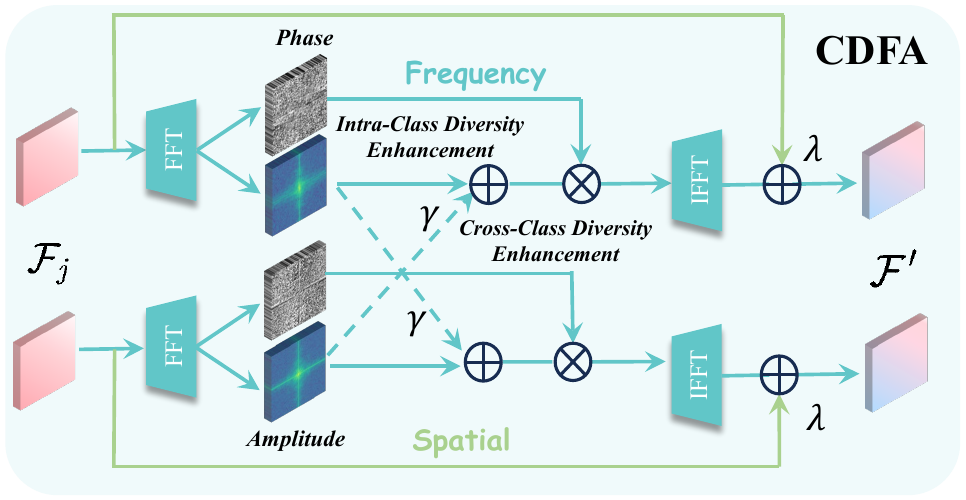}
    \caption{The proposed \textbf{CDFA} module introduces class diversity by leveraging the advantages of the frequency domain while preserving semantic details in the spatial domain.}
    \label{CDFA}
\end{figure}

\section{Methodology}
\label{Methodology}
\subsection{Motivation and Overview}
\textbf{\textit{KD} $\rightarrow$ \textit{DFKD}}. Given a training dataset \( D = \{(x_i, y_i)\}_{i=1}^{I} \), where each image \( x_i \in \mathbb{R}^{3 \times h \times w} \) and label \( y_i \in \{1,2,\dots,K\} \), \emph{Knowledge Distillation (KD)} transfers knowledge from a pre-trained teacher \( T(x;\theta_T) \) to a compact student \( S(x;\theta_S) \). The student learns to mimic the teacher’s output logits via
\begin{equation}
\begin{aligned}
\min_{\theta_S}\; &\mathbb{E}_{(x_i,y_i)\in D}\Big[
\KL\!\left(S(x_i;\theta_S)\,\|\,T(x_i;\theta_T)\right) \\
&\qquad\qquad
+ \lambda\,\CE\!\left(S(x_i;\theta_S),\,y_i\right)
\Big],
\end{aligned}
\label{eq:kd}
\end{equation}
where \( \lambda \) balances the Kullback--Leibler divergence and cross-entropy loss. By comparison, DFKD eliminates real data by adversarially training a generator \( G \) to synthesize samples for student learning:
\begin{equation}
\min_{\theta_S} \; \max_{\theta_G} \;
\mathbb{E}_{z \sim \mathcal{N}(0,1)} \Big[
\KL\!\big(S(\hat{x};\theta_S),\, T(\hat{x};\theta_T)\big)
\Big],
\label{eq:dfkd}
\end{equation}
where \( \hat{x} = G(z, y; \theta_G) \) denotes a synthetic image and \( z \) is sampled from a Gaussian prior.

\textbf{\textit{Motivation}.} Existing DFKD methods suffer from excessive reliance on teacher biases and frequency-specific dependencies, resulting in inconsistent synthetic image quality, limited class diversity, and unstable training. Moreover, current spatial-domain approaches remain overly abstract and fail to uncover the underlying frequency-domain causality behind these issues.

\textbf{\textit{CSWL-DFKD Framework}.} Figure~\ref{fig:overview} illustrates how the proposed method integrates seamlessly into the existing DFKD framework. During the generator update, the Class-Decoupled Frequency Augmentation (CDFA) module enhances class diversity by exploiting frequency-domain feature disentanglement, while the Cross-Stage Frequency Reconstruction (CSFR) auxiliary task stabilizes generator convergence ($\mathcal{L}_{G}^{R}$). Hence, the data generation stage jointly emphasizes \textbf{diversity} and \textbf{stability}. In the knowledge transfer stage, we apply explicit knowledge distillation (\(\mathcal{L}_{KL}\)) using these class-diverse synthetic images, and the CSFR module is reutilized to further improve the \textbf{generalization} and \textbf{representation transferability} of the student network (\(\mathcal{L}_{S}^{R}\)).

\subsection{Class-Decoupled Frequency Augmentation}
\label{3.2}
As shown in Figure~\ref{fig:Frequency Domain Analysis1}(a) and (b), existing generator-based DFKD methods suffer from strong dependence on the teacher’s class imbalance and prediction biases, leading to “shortcut learning” that exploits specific frequency components during optimization. To end this, we introduce the \underline{C}lass-\underline{D}ecoupled \underline{F}requency \underline{A}ugmentation (CDFA) module (Figure~\ref{CDFA}), which leverages frequency-domain representations to enhance class diversity while preserving spatial-domain semantics.

\textbf{Frequency Domain Representation.} The generative feature at the $j$-th layer of the generator is denoted by $F_j \in \mathbb{R}^{C \times H \times W}$, where $C$, $H$, and $W$ represent the number of channels, height, and width, respectively. For brevity, we may simply denote it as $F$. 
To convert the $F$ to the frequency space, we apply a two-dimensional Fast Fourier Transform (FFT) to each channel $c$, expressed as:
\begin{equation}
F_{\text{fft}}(c, k, l) = \sum_{h=0}^{H-1} \sum_{w=0}^{W-1} F(c, h, w) \cdot e^{-j2\pi \left(\frac{hk}{H} + \frac{wl}{W}\right)} 
\label{eq:1}
\end{equation}
where \( k \in [0, H-1] \) and \( l \in [0, W-1] \) denote the frequency positions. 
Then, the frequency domain generative representation is decomposed into amplitude and phase components:
\begin{equation}
F_{\text{fft}}(c, k, l) = A(c, k, l) \cdot e^{j\theta(c, k, l)} 
\end{equation}
where the amplitude $A(c, k, l)=\left|F_{\text{fft}}(c, k, l)\right|$ denotes the frequency intensity, and the phase $\theta(c, k, l)$ encodes positional information.

\textbf{Intra-Class Stochastic Amplitude Scaling.} 
To mitigate the negative effects of generative shortcut learning, we introduce a stochastic scaling method by applying the random scaling factor $\alpha_{k,l}$ to each frequency component $(k, l)$, seeks to disrupt the generator preference for low-level frequency patterns while reducing the risk of overfitting specific frequency components.
The scaled amplitude is defined as:
\begin{equation}
A'(c, k, l) = \alpha_{k,l} \cdot A(c, k, l) 
\end{equation}
where the scaling factor $\alpha_{k,l}$ is sampled from a uniform distribution $U(\alpha_{\text{min}}, \alpha_{\text{max}})$, with $\alpha_{\text{min}}$ and $\alpha_{\text{max}}$ controlling the range. 

\textbf{Inter-Class Amplitude Mixing.} 
Let \(F^{m} \in \mathbb{R}^{C \times H \times W} \) and \( F^{n} \in \mathbb{R}^{C \times H \times W} \) represent the frequency-domain features from category-$m$ and category-$n$, respectively. After applying stochastic amplitude scaling, their amplitudes are denoted as \( A'_m(c, k, l) \) and \( A'_n(c, k, l) \). The mixed amplitude is calculated as follows:
\begin{equation}
A_{\text{mixed}}(c, k, l) = A'_m(c, k, l) + A'_n(c, k, l), 
\end{equation}
Then, to combine the mixed amplitude \( A_{\text{mixed}}(c, k, l) \) and the native phase information \( \theta(c, k, l) \), the updated frequency domain feature is reconstructed by $A_{\text{mixed}}(c, k, l) \cdot e^{j\theta(c, k, l)}$, where \( \theta(c, k, l) \) represents the phase information, ensuring structural consistency while introducing inter-class diversity interactions. 
Finally, the class-enriched frequency features are reconstructed into the spatial domain using the inverse Fast Fourier Transform (IFFT), defined as:
\begin{equation}
\begin{aligned}
F'(c, h, w) = \frac{1}{HW} \sum_{k=0}^{H-1} \sum_{l=0}^{W-1} & A_{\text{mixed}}(c, k, l) \cdot e^{j\theta(c, k, l)} \\
& \cdot e^{j2\pi\left(\frac{kh}{H} + \frac{lw}{W}\right)}.
\end{aligned}
\end{equation}


\textbf{Generation with Augmented Features.} The proposed CDFA strategy jointly exploits the advantages of both the frequency and spatial domains to introduce class diversity and capture local structural details to the generative feature $F_{\text{aug}}$, which involves a weighted combination of the frequency-domain augmented feature $F'(c)$ and the native spatial feature $F(c)$:
\begin{equation}
\begin{aligned}
F_{\text{aug}}(c, h, w) = \lambda F'(c, h, w) + (1 - \lambda) F(c, h, w), \quad
\end{aligned}
\end{equation}
where  \( \lambda \) is a fusion coefficient controlling the balance between the two domains. The augmented feature $F_{\text{aug}}$ is then fed to the subsequent layers of the generator for the final image synthesis process:
\begin{equation}
\hat{x}_{aug} = G_{j+1}(F_{\text{aug}}) 
\label{eq:7}
\end{equation}
where \( G_{j+1} \) represents the remaining layers of the generator, starting from layer \(j+1\) to the last layer. 

\subsection{Cross-Stage Frequency Reconstruction}
\label{Cross-Stage Frequency Reconstruction}

While the proposed CDFA strategy introduces category diversity to synthetic images, it is acknowledged that the quality improvement is not stable, as shown in Section~\ref{Introduction} (Figure~\ref{fig:Frequency Domain Analysis1} (d)). We attribute this to two key factors: 
\textit{\textbf{i).}} The flat distribution required for diversity clashes with the sharp distribution needed for subsequent knowledge transfer \cite{han2021robustness}.
\textit{\textbf{ii).}}  Frequency-domain augmentation modifies spectral information, leading to semantic distortions and structural artifacts.
Inspired by the paradigm of improving training stability through regularization in  \cite{Metz2016}, we propose a Cross-Stage Frequency Reconstruction auxiliary task, which encourages focusing on essential generative constraints by reconstructing salient frequency information, thereby improving convergence stability. The specific optimization process is as follows.

\textbf{How Salient Frequency Information is Defined?} Previous studies have shown that the amplitude spectrum reflects an image’s energy distribution and highlights key visual features \cite{Chen2021,Zheng2021}. Based on this, we propose an amplitude-guided mechanism to generate challenging and informative mask regions. A synthetic image $\hat{x}$ uses a 2D Discrete Fourier Transform (DFT) represented as $F(\hat{x}) = A(\hat{x})e^{j\Phi(\hat{x})}$, where $A(\hat{x})$ is the amplitude spectrum and $\Phi(\hat{x})$ the phase spectrum. Larger amplitude frequency components indicate key visual information such as edges and textures. Thus, a binary mask $M(u, v)$ is constructed using the following rule:
\begin{equation}
M(u, v) =
\begin{cases} 
0, & \text{if } A(\hat{x})(u, v) > \tau, \\ 
1, & \text{otherwise},
\end{cases} 
\end{equation}
where $\tau$ is a mask threshold. To handle different object scales and complexities, $\tau$ is adaptive and computed from the distribution of $A(\hat{x})$ using the $p\%$ quantile, i.e., $\tau = Q_p(A(\hat{x})(u, v))$. Then, the mask $M(u, v)$ is applied to the frequency-domain representation of the image, resulting in a masked spectrum $F_m(\hat{x}) = M(u, v) \odot F(\hat{x})$, which is then transformed back to the spatial domain via inverse DFT (IDFT), yielding the masked image $\hat{x}_m = \mathcal{F}^{-1}(F_m(\hat{x}))$.

\textbf{Cross-Stage Semantic Reconstruction.} As discussed in Section \ref{Cross-Stage Frequency Reconstruction}, the instability factor-\textit{\textbf{i}} highlights the inconsistency between the optimization objectives during data generation and knowledge transfer phases. To mitigate semantic conflicts, we propose a shared frequency domain reconstruction objective. Specifically, the teacher network~($E_T$) in the data generation phase and the student network~($E_S$) in the knowledge transfer phase serve as encoders, but they share a decoder $ D_{shared}$ to provide training stability for the generator and improve generalization and representation transferability for the student network. During the update generator phase, reconstruction loss is defined as:
\begin{equation}
\mathcal{L}_{G}^{R} = \| D_{\text{shared}\text{-}\textcolor{blue}{Frozen}}({E_T(\hat{x}_m)}) - \hat{x} \|_2^2 
\label{eq:lgr}
\end{equation}
where $\hat{x}_m$ is the masked image, the \( \hat{x} \) denotes the input image, and $D_{\text{shared}\text{-}Frozen}$ denotes the shared decoder (D) whose weights remain frozen during the data generation phase. Similarly, during the knowledge transfer phase, reconstruction loss is defined as:
\begin{equation}
\mathcal{L}_{S}^{R} =  \| D_{\text{shared}\text{-}\textcolor{red}{Updatable}}({E_S(\hat{x}_m^{aug})}) - \hat{x}_{aug} \|_2^2 
\label{eq:sr}
\end{equation}
where $\hat{x}_{aug}$ originates from Equation~\ref{eq:7}, while $\hat{x}_m^{aug}$ represents the corresponding masked synthetic image. 

\textbf{Why Cross-Stage is Work?} 
During knowledge transfer, the decoder \(D\) is optimized with diversity enhanced synthetic samples \(\hat{x}^{aug}\) under the reconstruction loss \(L_{S}^{R}\). Benefiting from richer class diversity and data distributions, \(D\) learns a more robust reconstruction mapping. During data generation, the same decoder is shared with fixed parameters and operates in inference mode on single class semantic inputs. This design transfers the stable reconstruction capability learned in the knowledge transfer stage to the data generation stage, thereby alleviating training instability under single class constraints. By sharing \(D\) across the two stages, the method connects diverse semantic modeling with single class generation and implicitly forms a shared decoder with an exponential moving average characteristic, which further improves the stability and consistency of the overall training process \cite{alag2023ema,yaz2018unusual}.

\begin{algorithm}[!t]
   \caption{CSWL-DFKD with Shared Decoder}
   \label{alg:cswl}
\begin{algorithmic}[1]
   \STATE {\bfseries Input:} Teacher network $\mathcal{E}_T(\cdot;\theta_T)$, Student network $\mathcal{E}_S(\cdot;\theta_S)$, Generator $\mathcal{G}(\cdot;\theta_G)$, Shared decoder $\mathcal{D}(\cdot;\theta_D)$, Noise vector $z \sim \mathcal{N}(0, I)$, Epochs $E$, Iterations $n_G$, $n_S$
   \STATE {\bfseries Output:} Optimized student network $\mathcal{E}_S(\cdot;\theta_S)$

   \FOR{epoch $= 1$ \textbf{to} $E$}
      \STATE \textbf{// Stage 1: Data Synthesis (Update Generator)}
      \FOR{iter $= 1$ \textbf{to} $n_G$}
         \STATE $\hat{x} \leftarrow \mathcal{G}(z; \theta_G)$
         \STATE $\hat{x}_m \leftarrow \text{AmplitudeMask}(\hat{x})$ 
         \STATE Reconstruct masked image via frozen decoder: $\hat{x}_r \leftarrow \mathcal{D}_{\text{Frozen}}(\mathcal{E}_T(\hat{x}_m))$ 
         \STATE $\mathcal{L}_R^G \leftarrow \|\hat{x}_r - \hat{x}\|_2^2$
         \STATE Total Generator loss: $\mathcal{L}_G \leftarrow \mathcal{L}_{\text{cls}} + \mathcal{L}_{\text{BN}} + \mathcal{L}_R^G$ \COMMENT{Eq.~\ref{eq:generator_loss}}
         \STATE $\theta_G \leftarrow \theta_G - \eta_G \nabla_{\theta_G} \mathcal{L}_G$
      \ENDFOR

      \STATE \textbf{// Stage 2: Knowledge Transfer (Update Student Network)}
      \FOR{iter $= 1$ \textbf{to} $n_S$}
         \STATE Generate intermediate features: $F \leftarrow \mathcal{G}_j(z; \theta_G)$
         \STATE Augment features via CDFA: $F_{\text{aug}} \leftarrow \text{CDFA}(F)$
         \STATE $\hat{x}_{\text{aug}} \leftarrow \mathcal{G}_{j+1}(F_{\text{aug}})$ 
         \STATE $\hat{x}_{\text{aug},m} \leftarrow \text{AmplitudeMask}(\hat{x}_{\text{aug}})$
         \STATE Reconstruct masked image via updatable decoder: $\hat{x}_{\text{rec}} \leftarrow \mathcal{D}_{\text{Updatable}}(\mathcal{E}_S(\hat{x}_{\text{aug},m}))$ 
         \STATE $\mathcal{L}_R^S \leftarrow \|\hat{x}_{\text{rec}} - \hat{x}_{\text{aug}}\|_2^2$
         \STATE Total Student loss: $\mathcal{L}_S \leftarrow  \mathcal{L}_{\text{KL}} + \mathcal{L}_R^S + \mathcal{L}_{\text{CAM}}$ \COMMENT{Eq.~\ref{eq:student_loss}}
         \STATE $\theta_S \leftarrow \theta_S - \eta_S \nabla_{\theta_S} \mathcal{L}_S$
      \ENDFOR
   \ENDFOR
\end{algorithmic}
\end{algorithm}

\subsection{Optimization}

The optimization of the CSWL framework consists of two stages: generator updating and student network updating. The former corresponds to the Data Synthesis stage, which aims to generate synthetic samples with high diversity and discriminative capability. The latter corresponds to the Knowledge Transfer stage, which aims to achieve effective knowledge transfer from the teacher network to the student network using synthetic data. The pseudocode of the overall optimization procedure is presented in Algorithm \ref{alg:cswl}.

\textbf{Update Generator.} As illustrated in Figure~\ref{fig:overview}(a), the generator is optimized to produce diverse and meaningful synthetic data through the following objective function:
\begin{equation}
    \mathcal{L}_{{G}} = \mathcal{L}_{\mathrm{cls}} + \mathcal{L}_{\mathrm{BN}} + \mathcal{L}_{G}^{R}, 
    \label{eq:generator_loss}
\end{equation}
where $\mathcal{L}_{\mathrm{cls}}$ represents the cross-entropy loss, $\mathcal{L}_{\mathrm{BN}}$ is the batch normalization statistics alignment loss commonly used in DFKD, and $\mathcal{L}_{G}^{R}$ (defined in Equation~\ref{eq:lgr}) is the proposed cross-stage reconstruction loss that improves generator training stability.

\textbf{Update Student Network.} As shown in Figure~\ref{fig:overview} (a), the overall loss function for the student network is defined as:
\begin{equation}
    \mathcal{L}_S = \mathcal{L}_{\mathrm{KL}} + \mathcal{L}_{S}^{R} + \mathcal{L}_{\mathrm{CAM}},
    \label{eq:student_loss}
\end{equation}
where $\mathcal{L}_{\mathrm{KL}}$ denotes the Kullback–Leibler divergence, minimizing the prediction discrepancy between the teacher and student networks on the synthetic dataset. $\mathcal{L}_{\mathrm{CAM}}$ represents the activation region consistency loss \cite{Yu2023}.
$\mathcal{L}_{S}^{R}$ (defined in Equation~\ref{eq:sr}) is the proposed cross-stage reconstruction loss, which improves the student network’s generalizability and representational transferability during knowledge transfer.

\begin{table*}[!t]
\centering
\caption{\textbf{The comparison of distillation results of various methods on the multiple datasets.} The best-performing methods are highlighted in \textbf{bold}, while the runner-up methods are \underline{underlined}. Additionally, superscripts are used to indicate the sources of the results: \( {}^{\alpha}\) for  \cite{Tran2024}, \({}^{\beta}\) for  \cite{Li2024CCL}, and \({}^{\gamma}\) for  \cite{Yu2023}. In this table, "R" denotes ResNet, "W" represents WideResNet, and "V" stands for VGG. The higher accuracy between the teacher and student models is considered the upper bound in performance evaluation.}
\resizebox{\textwidth}{!}{
\begin{tabular}{c c c c c c c c c c c c c}
\toprule
\multirow{3}{*}{\textbf{Method}} & \multicolumn{5}{c}{\textbf{CIFAR10}} & \multicolumn{5}{c}{\textbf{CIFAR100}} & \textbf{TinyImageNet} & \textbf{ImageNet} \\ 
\cmidrule(lr){2-6} \cmidrule(lr){7-11}\cmidrule(lr){12-12} \cmidrule(lr){13-13}
 & R34 & W402 & W402 & W402 & V11 & R34 & W402 & W402 & W402 & V11 & R34 & R50 \\
 & R18 & W162 & W161 & W401 & R18 & R18 & W162 & W161 & W401 & R18 & R18 & R50 \\
\midrule
Teacher & 95.70 & 94.87 & 94.87 & 94.87 & 92.25 & 77.94 & 77.83 & 75.83 & 75.83 & 71.32 & 66.44 & 75.45 \\
Student & 95.20 & 93.95 & 91.12 & 93.94 & 95.20 & 77.10 & 73.56 & 65.31 & 72.19 & 77.10 & 64.87 & 75.45 \\
\midrule
\(\mathit{DeepInv}^\alpha\text{\cite{Micaelli2019}}\) & 93.26 & 89.72 & 83.04 & 86.85 & 90.36 & 61.32 & 61.34 & 53.77 & 68.58 & 54.13 & - & 68.00 \\
\(\mathit{ZSKT}^\alpha\text{\cite{Micaelli2019}}\) & 93.32 & 89.66 & 83.74 & 86.07 & 89.46 & 67.74 & 54.59 & 36.60 & 53.60 & 54.31 & - & - \\
\(\mathit{DFQ}^\alpha\text{\cite{Choi2020}}\) & 94.61 & 92.01 & 86.14 & 91.69 & 90.84 & 77.01 & 64.79 & 51.27 & 54.43 & 66.21 & - & - \\
\(\mathit{CMI}^\alpha\text{\cite{Fang2021}}\) & 94.84 & 92.52 & 90.01 & 92.78 & 91.13 & 77.04 & 68.75 & 57.91 & 68.88 & 70.56 & - & - \\
\(\mathit{IFHE}^\gamma\text{\cite{DAC}}\) & 95.09 & 93.59 & 91.80 & 93.71 & 92.01 & 77.11 & 70.61 & 61.55 & 69.95 & 70.98 & - & - \\
\(\mathit{SpaceshipNet}^\gamma\text{\cite{Yu2023}}\) & \underline{95.39} & 93.25 & 90.38 & 93.56 & 92.27 & 77.41 & 69.95 & 58.06 & 68.78 & 71.41 & 64.04 & - \\
\(\mathit{CCL-D}^\beta\text{\cite{Li2024CCL}}\) & 95.36 & 94.05 & 91.76 & \underline{94.27} & 91.58 & 77.12 & 70.59 & 61.04 & 70.21 & 71.02 & - & - \\
\(\mathit{NAYER}^\alpha\text{\cite{Tran2024}}\)
& 95.21 & \underline{94.07} & \underline{91.94} & 94.15 & \underline{92.37} & \underline{77.54} & \underline{71.72} & \underline{62.23} & \underline{71.80} & \underline{71.75} & \underline{64.17} & \underline{68.92} \\
\rowcolor{lightblue}
\textbf{Ours}  & \textbf{95.52} & \textbf{94.42} & \textbf{92.16} & \textbf{94.58} & \textbf{92.70} & \textbf{77.70} & \textbf{71.85} & \textbf{63.75} & \textbf{72.13} & \textbf{72.16} & \textbf{64.43} & \textbf{69.34} \\
\bottomrule
\end{tabular}}
\label{tab:sota_dfkd}
\end{table*}

\section{Experiments}
\label{Experiments}

\subsection{Experimental Settings}
\label{Experimental Settings}

\textbf{\textit{Datasets.}} We conduct a comprehensive evaluation of the proposed CSWL method on classification datasets with different input resolutions. For low resolution inputs of \(32 \times 32\), we evaluate CSWL on CIFAR-10 and CIFAR-100 \cite{Krizhevsky2009}, which contain 10 and 100 categories, respectively. For medium resolution inputs of \(64 \times 64\), we use Tiny-ImageNet \cite{Le2015}, which consists of 200 categories. For high resolution inputs of \(224 \times 224\), we perform evaluation on ImageNet-1K \cite{deng2009imagenet}. In addition, to further verify the transferability of the proposed method, we evaluate its performance on downstream tasks using NYUv2 \cite{Tran2024}.

\textbf{\textit{Configuration Details.}} We evaluated our approach using multiple backbone architectures \cite{Tran2024}, specifically ResNet \cite{He2016}, VGG \cite{Simonyan2015}, and WideResNet \cite{Zagoruyko2016}. During training, we employed the AdamW optimizer \cite{loshchilov2017decoupled} with a weight decay rate of 0.1. The learning rate for training the student network was set to 0.1 and decayed to 0 using a cosine annealing strategy \cite{loshchilov2016sgdr}. The generator was trained with a learning rate of 0.001, with 200 iterations per update. We set the batch size to 256 and conducted all experiments on NVIDIA A100 GPUs. Additionally, to optimize GPU memory usage, we applied automatic mixed precision during training \cite{Micikevicius2017}. For the experiments presented in Table~\ref{tab:sota_dfkd}, the models were trained for 300 epochs on the CIFAR-10 and CIFAR-100 datasets and for 400 epochs on the TinyImageNet dataset.

\subsection{Main Results}

\textbf{\textit{Comparisons with SOTA DFKD methods.}}
Table~\ref{tab:sota_dfkd} compares our method with several state-of-the-art DFKD approaches, including DeepInv \cite{Yin2020}, DFQ \cite{Choi2020}, ZSKT \cite{Micaelli2019}, CMI \cite{Fang2021}, CCL-D \cite{Li2024CCL}, SpaceshipNet \cite{Yu2023}, and NAYER \cite{Tran2024}. ``Teacher" and ``student" results indicate performance with access to the real dataset. Although SpaceshipNet \cite{Yu2023} improves the diversity of synthetic images in the spatial domain through feature diversity, its design remains relatively abstract and is still susceptible to noisy features introduced by random sampling. In contrast, CSWL exploits frequency domain modeling through CDFA to alleviate shortcut learning and enhance diversity, while CSFR further ensures semantic consistency and training stability. On CIFAR-10 and CIFAR-100 with an input resolution of \(32 \times 32\), as reported in Table~\ref{tab:sota_dfkd}, CSWL achieves higher accuracy across various teacher student pairs. In particular, on CIFAR-100 with WRN-40-2 as the teacher and WRN-16-1 as the student, CSWL outperforms SpaceShipNet by \(1.52\). On TinyImageNet with an input resolution of \(64 \times 64\), CSWL also consistently surpasses competing methods when ResNet-34 and ResNet-18 are used as the teacher and student, respectively. On ImageNet with an input resolution of \(224 \times 224\), CSWL likewise delivers competitive performance, achieving an improvement of \(0.42\).

\begin{table}[!t]
\centering
\caption{\textbf{Performance and Training Time Comparison on CIFAR-10 and CIFAR-100.} Accuracy (\%) and training time (hours) are reported using WRN40-2 as teacher and WRN16-2 as student on a single NVIDIA A100 GPU.}
\resizebox{\columnwidth}{!}{
\begin{tabular}{lcccc}
\toprule
\textbf{Dataset} & \textbf{DeepInv \cite{Yin2020}} & \textbf{CMI \cite{Fang2021}} & \textbf{SpaceshipNet \cite{Yu2023}} & \textbf{Ours} \\
\midrule
\multirow{2}{*}{CIFAR-10} & 12.06GB & 7.23GB & 5.72GB & \textbf{5.74GB} \\
& (31.23h) & (24.01h) & (22.35h) & (22.47h) \\
\midrule
\multirow{2}{*}{CIFAR-100} & 13.33GB & 8.53GB & 7.08GB  & \textbf{7.10GB} \\
& (31.23h) & (24.01h) & (22.37h) & (22.48h) \\
\midrule
\textbf{Average Speed Up} & 1$\times$ & 1.30$\times$ & 1.39$\times$ & \textbf{1.39$\times$} \\
\bottomrule
\end{tabular}}
\label{cost}
\end{table}

\textbf{\textit{Training time and computational cost comparison.}} To further validate the effectiveness of the proposed CSWL in practical applications, we compare it in Table~\ref{cost} with several state of the art methods that emphasize parameter efficiency and low computational cost. The results show that CSWL consistently maintains an advantage without introducing additional computational time or memory overhead.

\begin{figure}[!t]
    \centering
    \includegraphics[width=\linewidth]{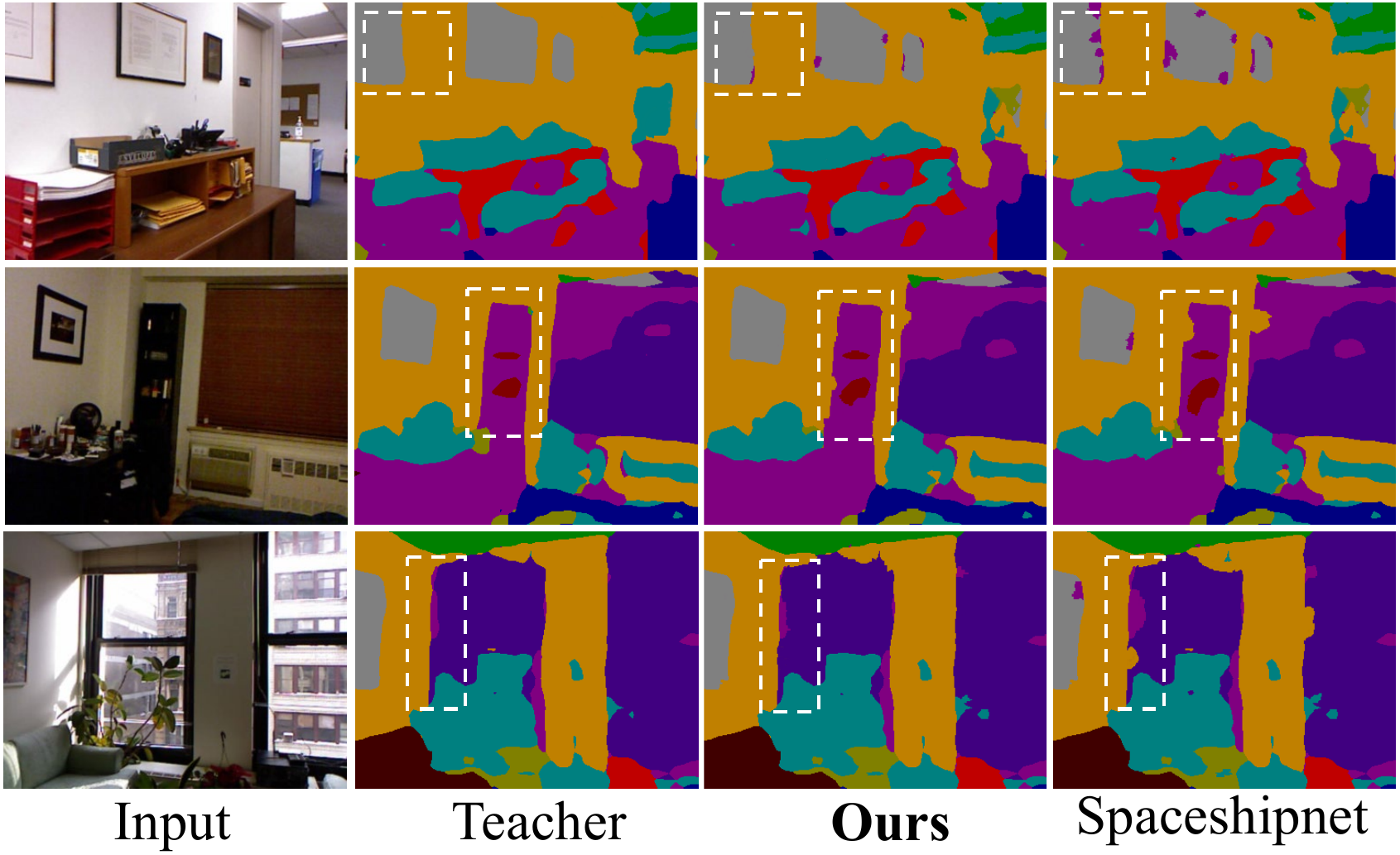}
    \caption{\textbf{Downstream Task Performance Visualization Comparison.} The output of the teacher network ResNet-34 is treated as the ground truth. Methods designed for optimization at the spatial level are less effective in improving the generalization ability of the student network in data free knowledge distillation \cite{Yu2023}. In contrast, our method learns domain-invariant class features, improving semantic segmentation and performance.}
    \label{fig:seg}
\end{figure}

\textbf{\textit{Transfer Learning Evaluation.}}
In addition to its strong performance in image recognition, our method also demonstrates remarkable transferability to downstream tasks. We conduct data free knowledge distillation on CIFAR-100, using ResNet-34 and ResNet-18 as the teacher and student networks, respectively. For the downstream semantic segmentation task, we adopt DeepLabV3 \cite{Chen2017} as the model architecture and evaluate performance on the NYUv2 dataset \cite{Silberman2012} using mIoU and pixAcc. As shown in Table~\ref{tabnyu}, our method consistently outperforms the current state of the art and achieves a 3.03 improvement in mIoU for semantic segmentation. In addition, Figure~\ref{fig:seg} presents qualitative visualization results of semantic segmentation.

\textbf{Additional Experiments in Data-free Quantization.} To further validate the generalizability of our data-free generation approach across different data-free learning scenarios, we extend its application to the task of data-free quantization. Specifically, We conducted a comparative analysis against ZeroQ \cite{cai2020zeroq}, AdaDFQ \cite{qian2023adaptive}, and HAST \cite{li2023hard}. As shown in Table~\ref{DFQ}, In this comparison, our method consistently achieved superior accuracy even in the data-free quantization setting, demonstrating strong potential for generalization to broader data-free applications.




\begin{table}[!t]
    \centering
    \caption{\textbf{The performance of the DeepLabv3 model on the NYUv2.} ResNet-34 and ResNet-18 are used as the teacher and student models, respectively.}
    \resizebox{\columnwidth}{!}{  
    \begin{tabular}{ccccc}
        \toprule
        \multirow{2}{*}{\textbf{Method}} & \multirow{2}{*}{\textbf{Data Available}} & \multirow{2}{*}{\textbf{Accuracy (\%)}} & \multicolumn{2}{c}{\textbf{NYUv2}} \\ 
        \cmidrule(l){4-5} 
        & & & \textbf{pAcc (\%)} & \textbf{mIoU} \\ 
        \midrule
        Teacher & \makecell{\centering \ding{51}} & 77.94 & 68.41 & 45.52 \\
        Student & \makecell{\centering \ding{51}} & 77.10 & 66.95 & 34.21 \\
        SpaceShipNet &  \ding{55} & 77.41 & 67.22 & 37.34  \\
        NAYER &  \ding{55} & 77.54 & 67.73 & 38.70  \\
        \rowcolor{lightblue} 
        \textbf{Ours} & \ding{55} & \textbf{77.70} & \textbf{68.24} & \textbf{41.73}  \\
        \bottomrule
    \end{tabular}
    }
    \label{tabnyu}
\end{table}

\begin{table}[!t]
\centering
\caption{\textbf{Quantization accuracy (\%) under W4A4 and W3A3 settings.} Results are reported on CIFAR-10 and CIFAR-100 with ZeroQ, AdaDFQ, HAST, and our method.}
\label{tab:quantization_comparison}
\resizebox{\columnwidth}{!}{
\begin{tabular}{lccccc}
    \toprule
    \textbf{Dataset} & \textbf{Bit} & \textbf{ZeroQ} \cite{cai2020zeroq} & \textbf{AdaDFQ} \cite{qian2023adaptive} & \textbf{HAST} \cite{li2023hard} & \textbf{Ours} \\
    \midrule
    \multirow{2}{*}{CIFAR-10} 
    & W4A4 & 84.68 & 92.31 & 92.36 & \textbf{93.21} \\
    & W3A3 & 29.32 & 84.89 & 88.34 & \textbf{89.16} \\
    \midrule
    \multirow{2}{*}{CIFAR-100} 
    & W4A4 & 58.42 & 66.81 & 66.68 & \textbf{66.92} \\
    & W3A3 & 15.38 & 52.74 & 55.67 & \textbf{56.32} \\
    \bottomrule
\end{tabular}
}
\label{DFQ}
\end{table}

\begin{table}[!t]
    \centering
    \caption{\textbf{Ablation study of different components.} Using VGG-11 as the teacher network and ResNet-18 as the student.}
    \resizebox{\columnwidth}{!}{ 
        \begin{tabular}{ccccc}
            \toprule
            \multirow{2}{*}{\bf Method} & \multicolumn{1}{c}{\bf Diversity} & \multicolumn{1}{c}{\bf Frequency} & \multicolumn{1}{c}{\bf Teacher: 71.32\%} \\
            \cmidrule(lr){4-4} 
            & \multicolumn{1}{c}{\bf method} & \multicolumn{1}{c}{\bf Reconstruction} & \multicolumn{1}{c}{\bf Student (\%)} \\ 
            \midrule
            SpaceshipNet & CFE & \ding{55} & 71.41 \\
            Ours w/o CSFR &  CDFA & \ding{55} &  71.83 \\
            Ours w/o  CDFA & CFE & \ding{51} & 71.95 \\
            \rowcolor{lightblue} 
            Ours &  CDFA & \ding{51} & {\bf 72.16} \\
            \bottomrule
        \end{tabular}
    }
    \label{tab:effectivenessCDFA}
\end{table}

\subsection{Ablation Study}


\hspace{1em}  \textbf{\textit{Effectiveness of CDFA.}} As discussed in Section~\ref{Introduction} (Figure~\ref{fig:Frequency Domain Analysis1}), generators tend to overfit specific frequency components, leading to overly simplified structures and suppressing the learning of diverse frequency semantics, which ultimately limits class diversity. To address this issue, we propose the Class-Decoupled Frequency Augmentation (CDFA) module, which performs frequency-domain decoupling to dynamically adjust the importance of different frequency bands, thereby mitigating class bias and enhancing semantic diversity. As shown in Table~\ref{tab:effectivenessCDFA}, without CDFA, the generator relies solely on spatial-domain features, resulting in degraded image quality and a noticeable drop in student network accuracy. Moreover, compared with the CFE based spatial feature enhancement adopted in SpaceshipNet \cite{Yu2023}, enhancement in the frequency domain is more effective at mitigating shortcut learning in the generator and further improves both class diversity and feature representation diversity.

We also conduct experiments to further evaluate the CDFA's impact by varying the random scaling factor \( \alpha \) and frequency domain feature mixing ratio \( \lambda \). The baseline uses a scaling factor \( \alpha \) within range \( (0.75, 1.8) \) and a mixing ratio \( \lambda \) of 0.8. Additional scaling ranges and mixing ratios are tested, while maintaining other settings from Section \ref{Experimental Settings}. Results are in Table~\ref{tab: CDFA_parameters_impact}. Excessively broad scaling factors $\alpha$ or high mixing ratios $\lambda$ introduce noise, reducing image quality.


\begin{table}[!t]
    \centering
    \caption{Performance comparison was evaluated on CIFAR-10 and CIFAR-100 using VGG-11 as the teacher and ResNet-18 as the student to evaluate the diversity of generated images. Higher IS and lower FID values indicate better quality and higher LPIPS means better diversity of the generated images.}
    \resizebox{\columnwidth}{!}{
    \begin{tabular}{llcccc}
        \toprule
        \textbf{Dataset} & \textbf{Approach} & \textbf{Accuracy} & \textbf{IS} & \textbf{FID} & \textbf{LPIPS} \\
        \midrule
        \multirow{4}{*}{CIFAR-10} 
        & NAYER & 92.37 & 1.61 & 318 & 0.14 \\
        & Spaceshipnet & 92.27 & 1.52 & 348 & 0.16 \\
        & ours w/o CSFR & 92.48 & 1.73 & 352 & 0.22 \\
        \rowcolor{lightblue} %
        & \textbf{ours} & \textbf{92.70} & \textbf{1.88} & \textbf{288} & \textbf{0.19} \\
        \midrule
        \multirow{4}{*}{CIFAR-100} 
        & NAYER & 71.75 & 1.83 & 345 & 0.22 \\
        & Spaceshipnet & 71.41 & 1.68 & 364 & 0.24 \\
        & ours w/o CSFR & 71.85 & 1.99 & 387 & 0.28 \\
        \rowcolor{lightblue} %
        & \textbf{ours} & \textbf{72.16} & \textbf{2.08} & \textbf{316} & \textbf{0.25} \\
        \bottomrule
    \end{tabular}
    }
    \label{tab:results}
\end{table}

\begin{table}[!t]
    \centering
    \caption{Analysis of the Impact of Random Scaling Factor \( \alpha \) and Frequency Domain Feature Mixing Ratio \( \lambda \) in  CDFA on the Performance Improvement of the Student Network.}
    \resizebox{\columnwidth}{!}{ 
        \begin{tabular}{ccccc}
            \toprule
            \multirow{2}{*}{\bf Method} & \multicolumn{4}{c}{\bf Random Scaling Factor \( \alpha \)} \\
            \cmidrule(lr){2-5} 
             & (0.5 , 2.0) & (0.6 , 1.5) & \textbf{(0.75 , 1.8)} & (0.9 , 1.2) \\
            \midrule
            {\bf Ours} & 71.13 & 71.58 & \textbf{72.06} & 71.62 \\
            \hline\hline
            \multirow{2}{*}{\bf Method} & \multicolumn{4}{c}{\bf Frequency Domain Feature Mixing Ratio \( \lambda \)} \\
            \cmidrule(lr){2-5} 
            & 0.3 & 0.5 & \textbf{0.8} & 0.9 \\
            \midrule
            {\bf Ours} & 71.54 & 71.73 & \textbf{72.06} & 71.82 \\
            \bottomrule
        \end{tabular}
    }
    \label{tab: CDFA_parameters_impact}
\end{table}

\begin{figure}[!t]
    \centering
    \includegraphics[width=\linewidth]{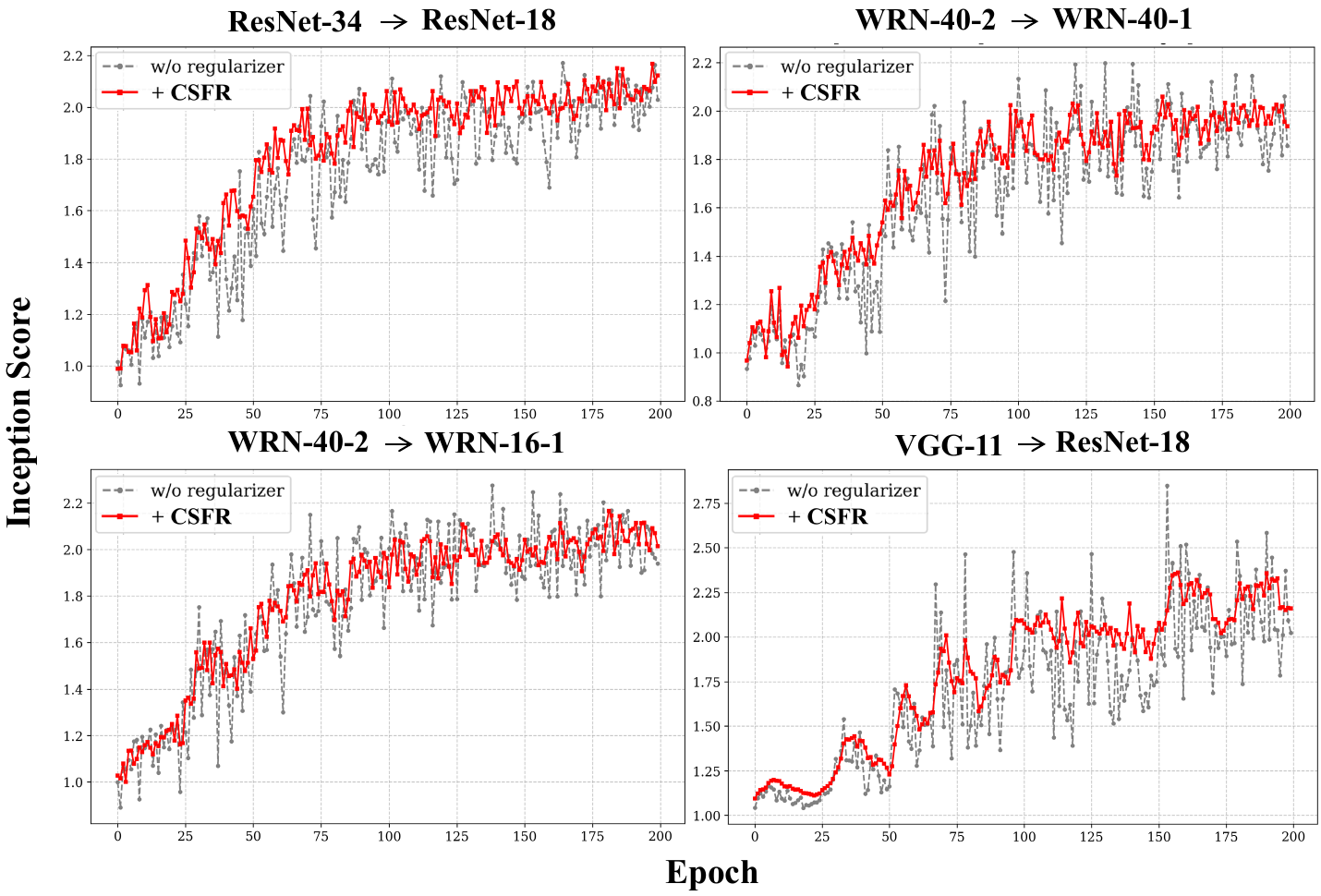}
    \caption{\textbf{The variation curve of Inception Score (IS) during the generator training process.} When CSFR is disabled, IS exhibits significant fluctuations, indicating instability in generator training. In contrast, enabling CSFR leads to a stable improvement in IS across all four teacher-student model combinations, validating its positive effect.}
    \label{fig:is_comp}
\end{figure}

\textbf{\textit{Effectiveness of CSFR.}} Although the frequency enhancement introduced by CDFA improves sample diversity, we observe that it also leads to training instability. This instability arises from the inherent conflict between the flat distribution required for class diversity and the sharp distribution needed for effective knowledge transfer. To address this issue, we propose a Cross Stage Frequency Reconstruction task, which enforces consistency in frequency domain features across different training stages. This regularization mitigates perturbation-induced fluctuations and promotes long-term stability and convergence. As shown in Figure~\ref{fig:is_comp}, disabling CSFR leads to instability after applying CDFA, while enabling it significantly stabilizes training. Table~\ref{tab:effectivenessCDFA} further provides quantitative evidence for the effectiveness of CSFR, showing that it yields superior synthetic image quality and diversity, as reflected by lower FID and higher LPIPS scores \cite{Zhang2018}.

\begin{figure*}[!]
    \centering 
    \includegraphics[width=\linewidth]{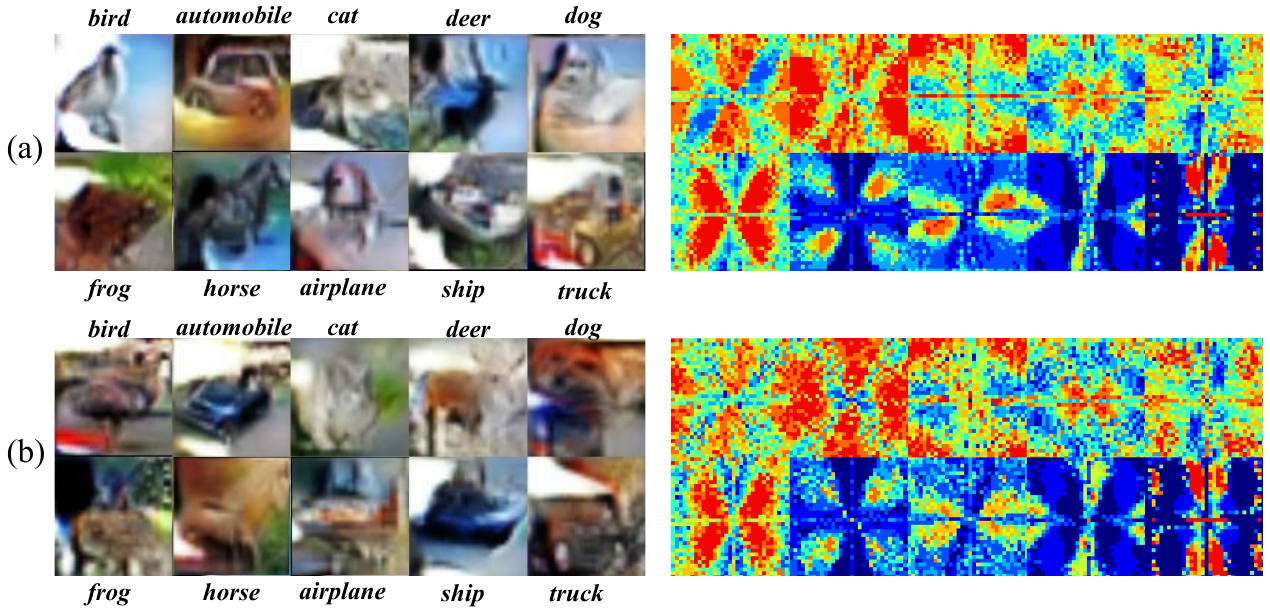} 
    \caption{\textbf{Synthesized Images from the Perspectives of the spatial Domain and Frequency Domain.} Panel (a) shows synthetic images generated in the spatial domain, while panel (b) presents those generated by our method from a frequency-domain perspective. In each panel, the left side displays the synthesized images corresponding to the CIFAR-10 dataset, and the right side shows their corresponding ADCS representations. It can be observed that our generated images exhibit more distinct ADCS patterns, indicating that CSWL enhances inter-class separability and intra-class consistency within the dataset.} 
    \label{fig:Frequency Domain Analysis} 
\end{figure*}

\begin{table}[!t] 
    \centering
    \caption{\textbf{Ablation experiments on the Effectiveness of Amplitude-Guided Masking} Evaluated on the CIFAR-100 dataset, with VGG-11 as the teacher network and ResNet-18 as the student network. Except for the masking method, all other aspects remain identical to the original approach.}
    \resizebox{0.85\linewidth}{!}{ 
        \begin{tabular}{cc}
            \toprule
            {\bf Method} & {\bf Student (\%)} \\ 
            \midrule
            Random Mask & 71.85 \\
            High-Frequency Mask & 71.62 \\
            Low-Frequency Mask & 71.50 \\
            Phase Gradient Mask & 71.87 \\
            \rowcolor{lightblue} 
            Amplitude-Guided Mask\textbf{(Ours)} & {\bf 72.06} \\
            \bottomrule
        \end{tabular}
    }
    \label{tab:effectiveness_mask}
\end{table}

\textbf{\textit{Effectiveness of Amplitude-Guided Masking.}} We conduct an ablation study on the Amplitude-Guided Masking method introduced in Section~\ref{Cross-Stage Frequency Reconstruction} and compare it with several alternative masking strategies to validate its effectiveness for frequency-domain reconstruction. Specifically, we evaluate four masking strategies and quantitatively analyze their impact, with the results reported in Table~\ref{tab:effectiveness_mask}. We observe that both High-Frequency Masking and Low-Frequency Masking lead to inferior performance. A plausible explanation is that masking a fixed frequency band forces the decoder to reconstruct missing components from a biased spectral pattern, which leads to suboptimal reconstruction. In contrast, Phase Gradient Masking may damage class-relevant semantic information, since phase encodes critical structural and semantic cues. By comparison, the amplitude spectrum primarily reflects the energy distribution across frequency components. Therefore, amplitude-guided masking is less likely to corrupt class semantics while still producing more challenging masked regions, making it particularly suitable for Cross-Stage Frequency Reconstruction. In addition, Masking ratio also plays an important role in model performance and generalization. To study its effect, we vary the quantile parameter \(p\) in Equation~10, which controls the proportion of high-amplitude frequency components covered by the mask. For example, \(p = 90\%\) masks the top \(10\%\) highest-amplitude components. As shown in Table~\ref{tab:p}, the best performance is achieved at \(p = 90\%\), which is therefore adopted as the default setting in all experiments.

\begin{table}[!t]
    \centering
    \caption{\textbf{Impact of different masking intensities on model performance.} Evaluated on the CIFAR-100 dataset, with VGG-11 as the teacher network and ResNet-18 as the student network. Except for the masking intensity, all other aspects remain identical to the original approach.}
    \resizebox{\linewidth}{!}{ 
        \begin{tabular}{ccccc}
            \toprule
            \multirow{2}{*}{\bf Method} & \multicolumn{4}{c}{\bf Quantile parameter \( p \)} \\
            \cmidrule(lr){2-5} 
             & \( p = 50\% \) & \( p = 80\% \) & \textbf{\( p = 90\% \)} & \( p = 95\% \) \\
            \midrule
            {\bf Amplitude-Guided Mask}  & 71.58 & 71.86 & \textbf{72.06} & 71.91 \\
            \bottomrule
        \end{tabular}
    }
    \label{tab:p}
\end{table}

\begin{figure}[!t]
    \centering
    \includegraphics[width=\linewidth]{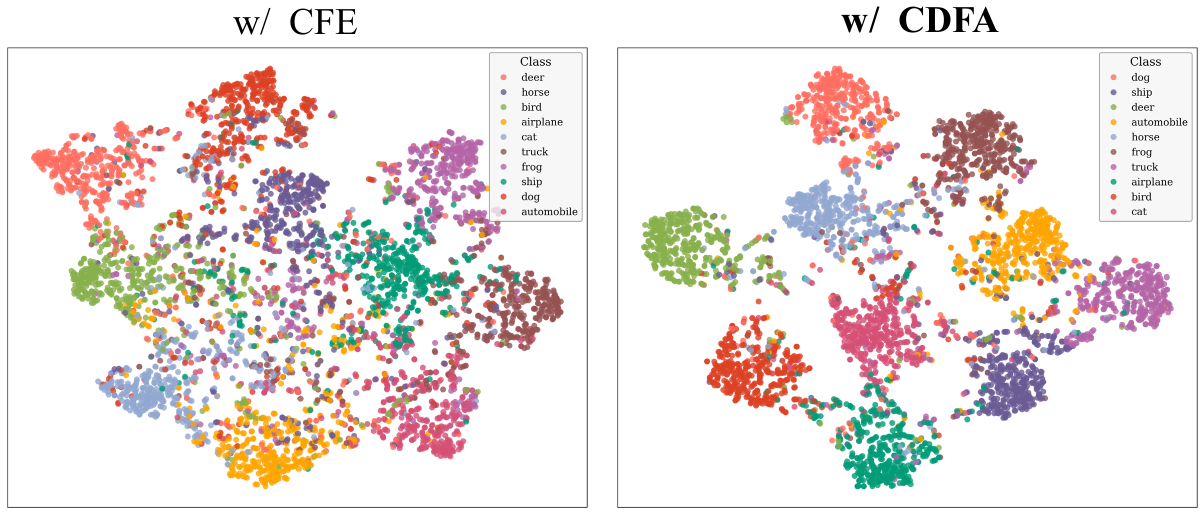}
    \caption{\textbf{Impact of  CDFA on Category Distribution.} CDFA provides a more compact intra-class distribution and clearer inter-class distances, thereby enhancing the quality of the data source used for knowledge distillation.}
    \label{t-SNE results on different datasets}
\end{figure}

\subsection{Visualization} 
We further visualize the feature distributions of real and synthetic images using t-SNE on CIFAR-10, with ResNet-34 as the feature extractor. Specifically, we sample 5,000 synthetic images from 200 batches for visualization. As shown in Figure~\ref{t-SNE results on different datasets}, although CFE \cite{Yu2023} improves image diversity through spatial feature exchange, it also blurs the decision boundaries of noisy classes, leading to greater class overlap and reduced classification accuracy. In contrast, CDFA improves the fidelity of synthetic samples through frequency-domain modeling and significantly enhances class separability, resulting in more compact intra-class distributions and clearer inter-class boundaries. These results suggest that CDFA not only increases sample diversity but also preserves more discriminative category semantics, thereby improving knowledge transfer in data-free knowledge distillation.

\textbf{\textit{Visually analyzing the quality of synthetic images from a frequency-domain perspective.}} We collected several synthesized images during the mid-training phase (epochs 150–199), as shown in Figure~\ref{fig:Frequency Domain Analysis}. From a spatial-domain perspective, at first glance, our method shows no significant difference in image quality or diversity compared to spatial-domain exploration methods. However, from a frequency-domain perspective, our approach enhances class diversity and consistency, as evidenced by more distinct ADCS \cite{Wang2023a} patterns across different classes. This distinction highlights inter-class differences within similar datasets, ultimately benefiting the student network, with quantitative validation provided in Table~\ref{tab:results}.

\section{Conclusion}
In this paper, we focus on two critical challenges in Data-Free Knowledge Distillation (DFKD): (i) the generator's over-reliance on teacher preferences and specific frequency patterns, which leads to insufficient class diversity and inconsistent quality in synthetic samples, and (ii) the instability of the training process. To address these issues, we propose the CSWL framework, which improves both diversity and stability in DFKD from a frequency-domain perspective.

Our contributions are mainly supported by two complementary components. First, we introduce the Class-Decoupled Frequency Augmentation (CDFA) module, which disentangles amplitude and phase information in generative features and incorporates inter-class augmentation in the frequency domain. This design reduces the generator's dependence on specific frequency components and positions, effectively suppresses generative shortcut learning, and improves the class diversity of synthetic images. Second, we propose the Cross-Stage Frequency Reconstruction (CSFR) task, which shares a frequency-domain reconstruction objective across the data generation and knowledge transfer stages. By enforcing consistency in the recovery of salient frequency information, CSFR introduces a stable regularization signal for generator optimization and further enhances the generalization and transferability of student representations.

Experimental results on multiple image recognition benchmarks with different input resolutions, and downstream tasks, show that CSWL consistently outperforms existing mainstream DFKD methods. Further analysis suggests that these gains arise from an effective combination of two mechanisms: frequency-domain augmentation improves class diversity by reducing the generator's reliance on local frequency patterns, while cross-stage reconstruction stabilizes optimization and improves synthetic data quality and training stability.

\textbf{Future Work:} Although this work improves DFKD from the perspectives of frequency-domain augmentation and cross-stage reconstruction, several important directions remain for future study. First, existing DFKD methods are still primarily developed for relatively standard settings. We plan to extend CSWL to support a broader range of more complex network architectures. Second, the current study is primarily centered on image classification. Although we have additionally validated the method on semantic segmentation and data-free quantization, it remains important to explore its applicability to more challenging vision tasks, such as object detection, instance segmentation, and multimodal scenarios, in order to assess its broader generality.

\bibliographystyle{IEEEtran}  
\bibliography{IEEEexample}




\end{document}